\documentclass[10pt,twocolumn,letterpaper]{article}

\usepackage{iccv}
\usepackage{times}
\usepackage{epsfig}
\usepackage{graphicx}
\usepackage{amsmath}
\usepackage{amssymb}
\usepackage{wrapfig}

\usepackage{amsmath,amsfonts,bm}

\def\1{\bm{1}}

\def\vh{{\bm{h}}}

\def\vm{{\bm{m}}}
\def\vn{{\bm{n}}}

\def\vv{{\bm{v}}}

\def\vx{{\bm{x}}}

\def\vz{{\bm{z}}}

\DeclareMathAlphabet{\mathsfit}{\encodingdefault}{\sfdefault}{m}{sl}
\SetMathAlphabet{\mathsfit}{bold}{\encodingdefault}{\sfdefault}{bx}{n}

\def\sR{{\mathbb{R}}}

\newcommand{\E}{\mathbb{E}}

\newcommand{\x}{$\times$}

\usepackage[super]{nth}
\usepackage[dvipsnames]{xcolor}
\usepackage{paralist, tabularx} %
\usepackage[caption=false, font=small]{subfig}
\usepackage{multirow}
\usepackage{dsfont}
\usepackage{transparent}
\usepackage{enumitem}
\usepackage{stmaryrd}
\usepackage{chngcntr}

\usepackage[pagebackref=true,breaklinks=true,letterpaper=true,colorlinks,bookmarks=false]{hyperref}

\iccvfinalcopy % *** Uncomment this line for the final submission

\def\iccvPaperID{10303} % *** Enter the ICCV Paper ID here

\ificcvfinal\fi

\newcommand{\dcs}{DetCon$_{S}$\ }
\newcommand{\dcb}{DetCon$_{B}$\ }
\newcommand{\apbbox}[1]{AP$^\text{bb}_\text{#1}$}
\newcommand{\apmask}[1]{AP$^\text{mk}_\text{#1}$}

\newlength\savewidth
  
\definecolor{Highlight}{HTML}{39b54a}  % green
\definecolor{Highlineg}{HTML}{B54A39}  % red

\newcolumntype{x}[1]{>{\centering\arraybackslash}p{#1pt}}
\newcolumntype{y}[1]{>{\raggedright\arraybackslash}p{#1pt}}
\newcolumntype{z}[1]{>{\raggedleft\arraybackslash}p{#1pt}}

\usepackage{tabularx,booktabs}
\newcolumntype{Y}{>{\centering\arraybackslash}X}

\begin{document}

\title{Efficient Visual Pretraining with Contrastive Detection}

\author{Olivier J. Hénaff  \quad Skanda Koppula \quad Jean-Baptiste Alayrac \\ Aaron van den Oord \quad Oriol Vinyals \quad João Carreira \vspace{.5em} \\
DeepMind, London, UK\\
}

\maketitle
\ificcvfinal\thispagestyle{empty}\fi

%%%%%%%%% ABSTRACT
\begin{abstract}

Self-supervised pretraining has been shown to yield powerful representations for transfer learning. These performance gains come at a large computational cost however, with state-of-the-art methods requiring an order of magnitude more computation than supervised pretraining. We tackle this computational bottleneck by introducing a new self-supervised objective, contrastive detection, which tasks representations with identifying object-level features across augmentations. This objective extracts a rich learning signal per image, leading to state-of-the-art transfer accuracy
on a variety of downstream tasks, while requiring up to 10\x \ less pretraining.
% from ImageNet to COCO, while requiring up to 5\x \ less pretraining. 
In particular, our strongest ImageNet-pretrained model performs on par with SEER, one of the largest self-supervised systems to date, which uses 1000\x \ more pretraining data. Finally, our objective seamlessly handles pretraining on more complex images such as those in COCO, closing the gap with supervised transfer learning from COCO to PASCAL. 
\end{abstract}

\vspace{-1em}

%%%%%%%%% BODY TEXT
\section{Introduction}

Since the AlexNet breakthrough on ImageNet, transfer learning from large labeled datasets has become the dominant paradigm in computer vision \cite{krizhevsky2012imagenet, russakovsky2015imagenet}. While recent advances in self-supervised learning have alleviated the dependency on labels for pretraining, they have done so at a tremendous computational cost, with state-of-the-art methods requiring an order of magnitude more computation than supervised pretraining \cite{caron2020unsupervised, chen2020big, grill2020bootstrap}. Yet the promise of self-supervised learning is to harness massive unlabeled datasets, making its computational cost a critical bottleneck.

\begin{figure}[t]
    \begin{center}
        \includegraphics[width=0.85\linewidth]{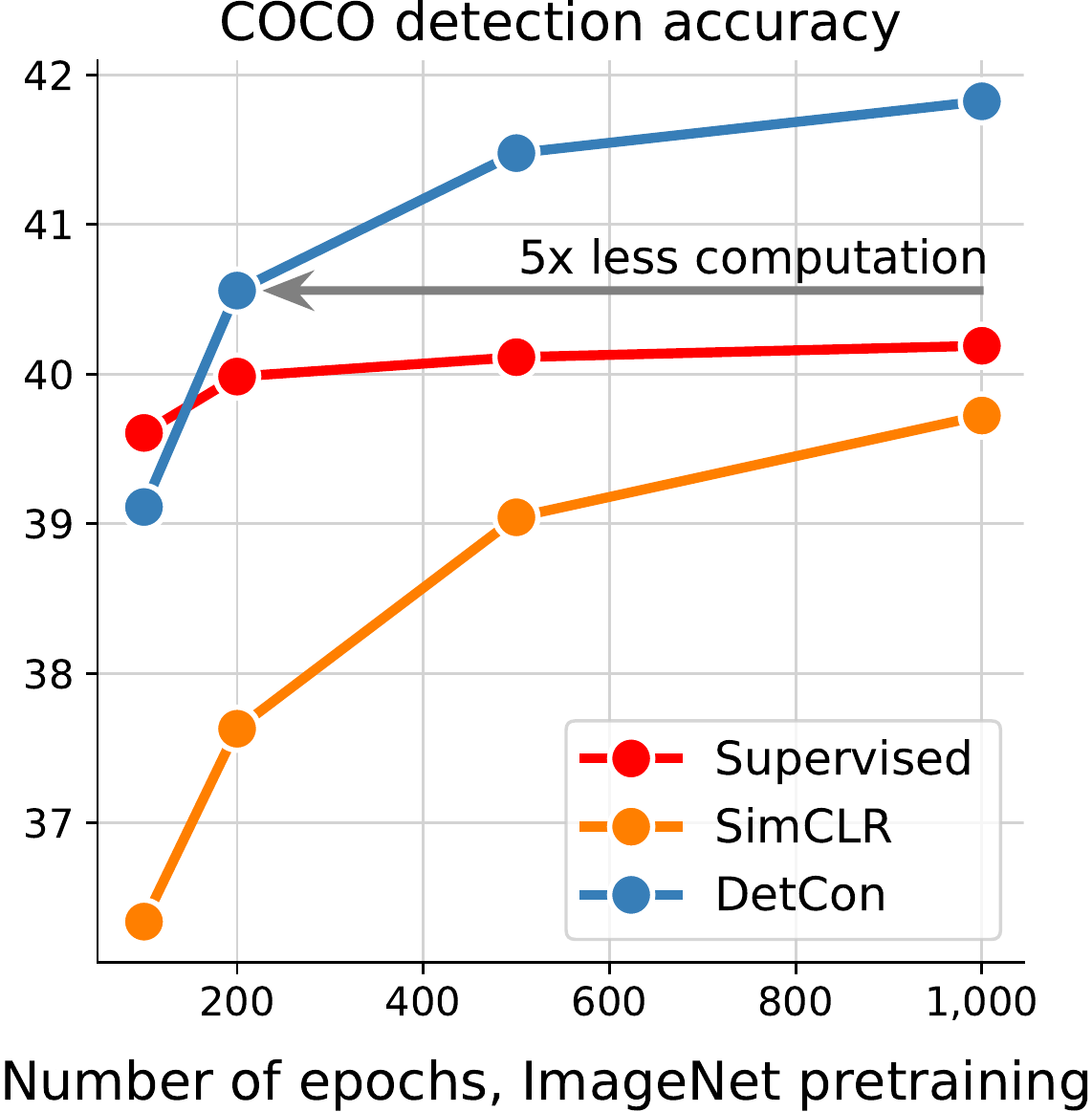}
    \end{center}
   \caption{\textbf{Efficient self-supervised pretraining with DetCon.} Self-supervised pretraining with SimCLR~\cite{chen2020simple} matches the transfer performance of supervised pretraining only when given 10\x \ more training iterations. Our proposed DetCon objective surpasses both, while requiring 5\x \ less computation than SimCLR. Transfer performance is measured by fine-tuning the representation on the COCO dataset for 12 epochs, using a Mask-RCNN.}
\label{fig:intro}
\vspace{-2.em}
\end{figure}

In this work, we aim to alleviate the computational burden of self-supervised pretraining.
To that end we introduce \textit{contrastive detection}, a new objective which maximizes the similarity of object-level features across augmentations. The benefits of this objective are threefold. First, it extracts separate learning signals from all objects in an image, enriching the information provided by each training example for free---object-level features are simply obtained from intermediate feature arrays. Second, it provides a larger and more diverse set of \textit{negative samples} to contrast against, which also accelerate learning. Finally, this objective is well suited to learning from complex scenes with many objects, a pretraining domain that has proven challenging for self-supervised methods. 

We identify approximate object-based regions in the image through the use of unsupervised segmentation algorithms. Perceptual grouping~\cite{koffka2013principles,martin2001database}---the idea that low and mid-level regularities in the data such as color, orientation and texture allow for approximately parsing a scene into connected surfaces or object parts---has long been theorized to be a powerful prior for vision~\cite{guerrero2008image, lowe2012perceptual, ullman1996high}. We leverage these priors by grouping local feature vectors accordingly, and applying our contrastive objective to each object-level feature separately. We investigate the use of several unsupervised, image-computable masks \cite{felzenszwalb2004efficient, arbelaez2014multiscale}, and find our objective to work well despite their inaccuracies.

We test the ability of our objective to quickly learn transferable representations by applying it to the ImageNet dataset and measuring its transfer performance on challenging tasks such as COCO detection and instance segmentation, semantic segmentation on PASCAL and Cityscapes, and NYU depth estimation. Compared to representations obtained from recent self-supervised %\textit{contrastive recognition} %instance discrimination 
objectives such as SimCLR and BYOL \cite{chen2020simple, grill2020bootstrap}, our representations are more accurate and can be obtained with much less training time. We also find this learning objective to better handle images of more complex scenes, bridging the gap with supervised transfer from the COCO dataset. In summary, we make the following contributions:  % 

\vspace{0.25em} 1. We formulate a new contrastive objective which maximizes the similarity across augmentations of all objects in a scene, where object regions are provided by a simple, unsupervised heuristic. %We dissect this new objective and assess the improvements afforded by each of its elements.

\vspace{0.25em} 2. We find this objective to alleviate the computational burden of self-supervised transfer learning, reducing by up to 10\x \ the computation required to match supervised transfer learning from ImageNet. Longer training schedules lead to state-of-the-art transfer to COCO detection and instance segmentation, and our best model matches the very recent state-of-the-art self-supervised system SEER~\cite{goyal2021self} which is trained on 1000\x \ more---if less curated---images.  %  (1 billion)

\vspace{0.25em} 3. When transferring from complex scene datasets such as COCO, our method closes the gap with a supervised model which learns from human-annotated segmentations. 

\vspace{0.25em} 4. Finally, we assess to what extent the existing contrastive learning paradigm could be simplified in the presence of high quality image segmentations, raising questions and opening avenues for future work. 

\section{Related work}

Transferring the knowledge contained in one task and dataset to solve other downstream tasks (i.e.\ \textit{transfer learning}) has proven very successful in a range of computer vision problems \cite{girshick2014rich, long2015fully}. While early work focused on improving the pretraining architecture \cite{he2016deep, simonyan2014very} and data \cite{sun2017revisiting}, recent work in self-supervised learning has focused on the choice of pretraining objective and task. 
Early self-supervised pretraining typically involved image restoration, including denoising \cite{vincent2008extracting}, inpainting \cite{pathak2016context}, colorization \cite{zhang2016colorful, larsson2017colorization}, egomotion prediction~\cite{agrawal2015learning}, and more \cite{donahue2016adversarial, nathan2018improvements, zhang2017split}. Higher-level pretext tasks have also been studied, such as predicting context \cite{doersch2015unsupervised}, orientation \cite{gidaris2018unsupervised}, spatial layouts \cite{noroozi2016unsupervised}, temporal ordering \cite{misra2016shuffle}, and cluster assignments \cite{caron2018deep}. 

Contrastive objectives, which maximize the similarity of a representation across views, while minimizing its similarity with distracting negative samples, have recently gained considerable traction \cite{hadsell2006dimensionality}. These views have been defined as local and global crops \cite{hjelm2018learning, bachman2019learning, oord2018representation, henaff2019data} or different input channels \cite{tian2019contrastive}. Instance-discrimination approaches generate global, stochastic views of an image through data-augmentation, and maximize their similarity relative to marginally sampled negatives \cite{chen2020simple, doersch2017multi, dosovitskiy2014discriminative, he2019momentum, wu2018unsupervised}, although the need  for negative samples has recently been questioned \cite{chen2020exploring, grill2020bootstrap}. While the benefits of instance-discrimination approaches have mostly been limited to pretraining from simple datasets such as ImageNet, clustering-based pretraining has proven very successful in leveraging large amounts of uncurated images for transfer learning \cite{asano2019self, caron2019leveraging, caron2020unsupervised, goyal2021self, ji2019invariant}.

While most work has focused on learning whole-image representations, there has been increasing interest in learning local descriptors that are more relevant for downstream tasks such as detection and segmentation. Examples of such work include the addition of auxiliary losses \cite{tian2020makes}, architectural components \cite{pinheiro2020unsupervised}, or both \cite{xie2020propagate}. While perceptual grouping has long been used for representation learning, often relying on coherent motion in videos \cite{li2016unsupervised, pathak2016learning, wang2015unsupervised}, it has only recently been combined with contrastive learning \cite{jabri2020space, van2021unsupervised, zhang2020self}. Most related to our work are \cite{van2021unsupervised, zhang2020self} that also leverage image segmentations for self-supervised learning, although both differ from ours in that they learn backbones that are specialized for semantic segmentation and employ different loss functions. Although these works arrive at impressive unsupervised segmentation accuracy, neither report gains in pretraining efficiency for transfer learning tasks such as COCO detection and instance segmentation, which we study next.

\section{Method}

\noindent We introduce a new contrastive objective which maximizes the similarity across views of local features which represent the same object (Figure \ref{fig:methods}). In order to isolate the benefit of these changes, we make the deliberate choice of re-using elements of existing contrastive learning frameworks where possible. To test the generality of our approach, we derive two variants, \dcs and DetCon$_{B}$, based on two recent self-supervised baselines, SimCLR \cite{chen2020simple} and BYOL \cite{grill2020bootstrap} respectively. We adopt the data augmentation procedure and network architecture from these methods while applying our proposed \textit{contrastive detection} loss to each.

\subsection{The contrastive detection framework}
\label{sec:method_framework}

\begin{figure}[t]
    \begin{center}
        \includegraphics[width=0.88\linewidth]{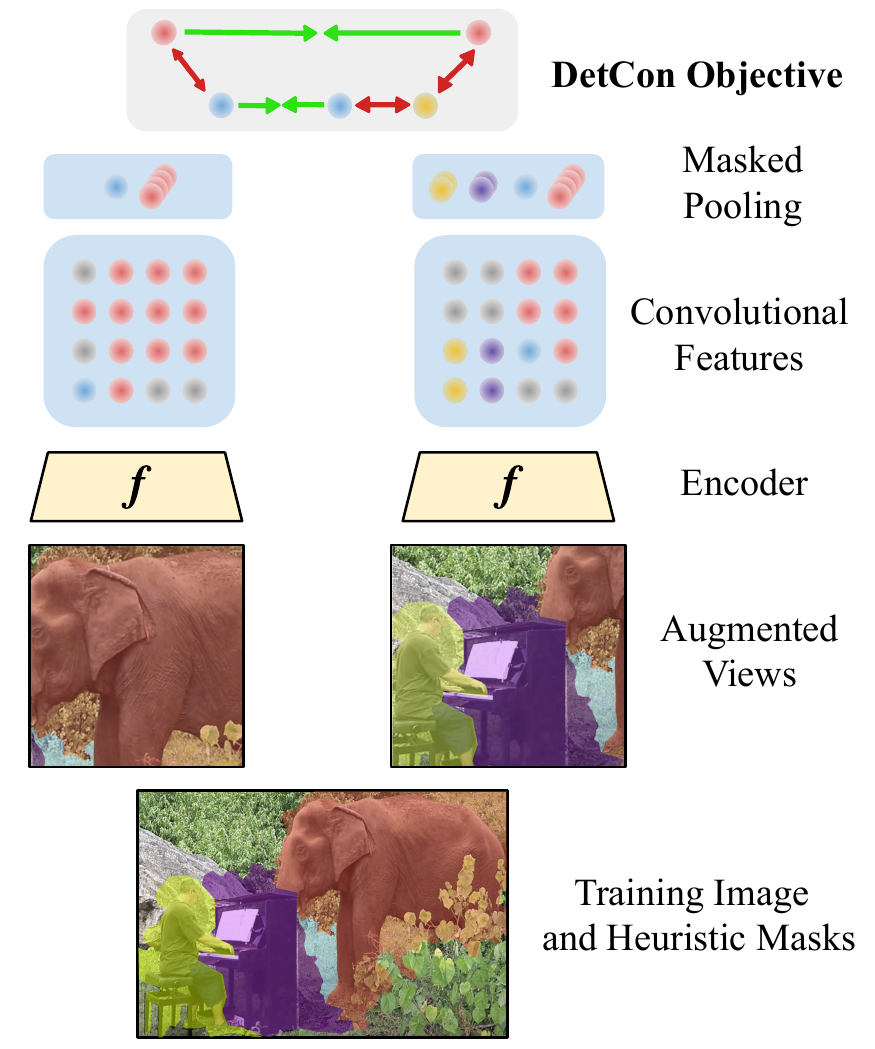}
    \end{center}
    \vspace{-1em}
  \caption{\textbf{The contrastive detection method.} We identify object-based regions with approximate, image-computable segmentation algorithms (bottom). These masks are carried through two stochastic data augmentations and a convolutional feature extractor, creating groups of feature vectors in each view (middle). The contrastive detection objective then pulls together pooled feature vectors from the same mask (across views) and pushes apart features from different masks and different images (top).}
\label{fig:methods}
\vspace{-0.5em}
\end{figure}

\vspace{0.5em} \noindent \textbf{Data augmentation.} Each image is randomly augmented twice, resulting in two images: $\vx, \vx'$. DetCon$_{S}$ and DetCon$_{B}$ adopt the augmentation pipelines of SimCLR and BYOL respectively, which roughly consist of random cropping, flipping, blurring, and point-wise color transformations. We refer the reader to appendix \ref{sec:app-data} for more details. In all cases, images are resized to 224$\times$224 pixel resolution.

In addition, we compute for each image a set of masks which segment the image into different components. As described in Section \ref{sec:method_masks}, these masks can be computed using efficient, off-the-shelf, unsupervised segmentation algorithms. If available, human-annotated segmentations can also be used. In any case, we transform each mask (represented as a binary image) %geometrically 
using the same cropping and resizing as used for the underlying RGB image, resulting in two sets of masks $\{\vm\}, \{\vm'\}$ which are aligned with the augmented images $\vx, \vx'$ (see Figure \ref{fig:methods}, \textit{augmented views}).

\begin{figure*}
  \begin{minipage}[c]{0.75\textwidth}
    \includegraphics[width=\textwidth]{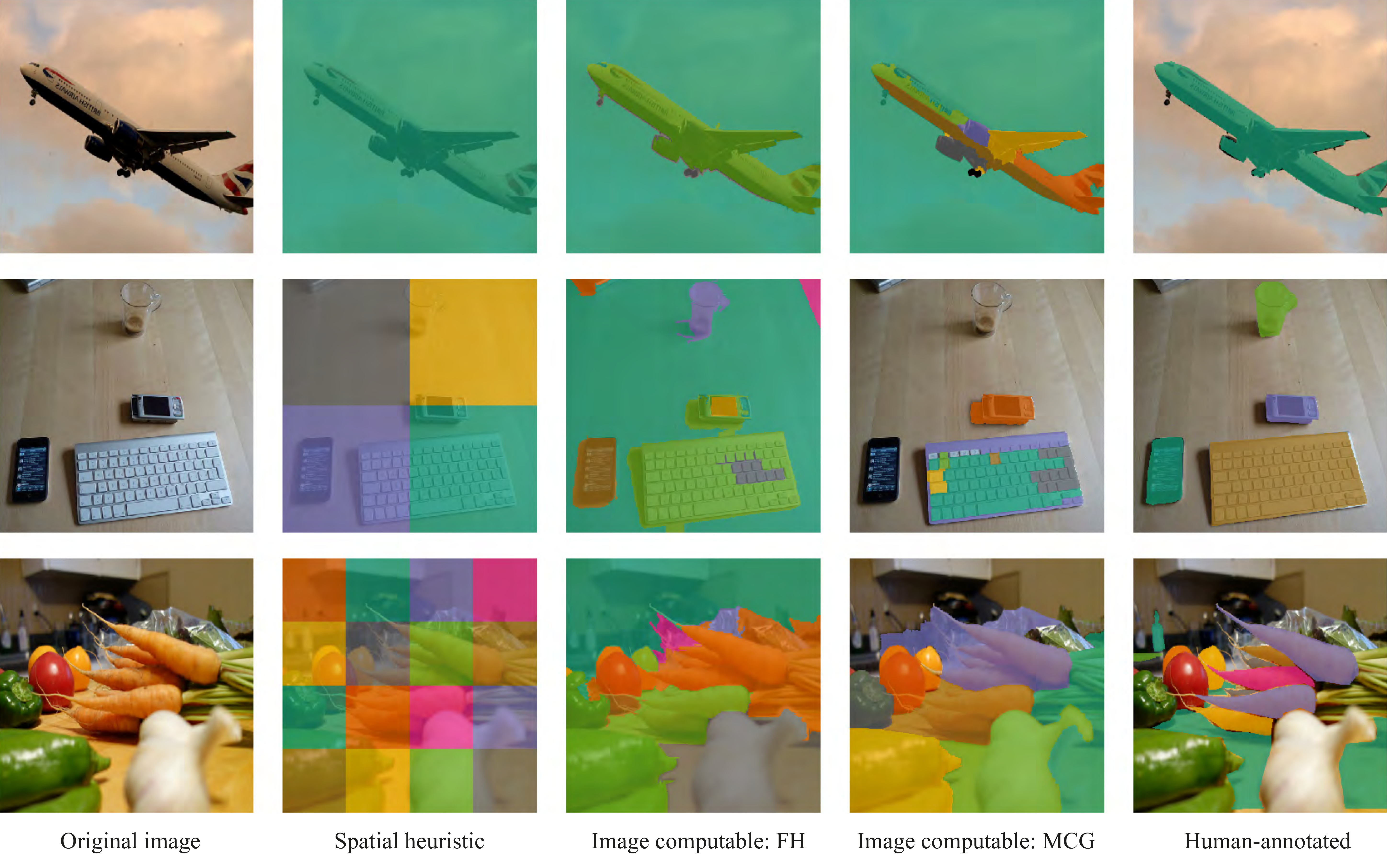}
  \end{minipage}\hfill
  \begin{minipage}[c]{0.23\textwidth}
  \caption{\textbf{Example masks used by the DetCon model.} \hspace{3em} \nth{1} column: random images from the COCO training set. \nth{2} column: masks based on spatial proximity only. Global masks (top) are implicitly used by methods such as SimCLR, MoCo, and BYOL. \nth{3} column: image-computable masks obtained from the Felzenszwalb-Huttenlocher (FH, \cite{felzenszwalb2004efficient}) algorithm, with $s=500$. \nth{4} column: image-computable masks inferred using Multiscale Combinatorial Grouping (MCG) \cite{arbelaez2014multiscale}. \hspace{3em} \nth{5} column: ``oracle'' masks used to assess potential improvements from higher-quality segmentations. \vspace{1.5em}} \label{fig:masks}
\end{minipage}
\vspace{-1em}
\end{figure*}

\vspace{0.5em} \noindent \textbf{Architecture.} We use a convolutional feature extractor $f$ to encode each image with a spatial map of hidden vectors: $\vh = f(\vx)$ where $\vh \in \sR^{H \times W \times D}$. We use the output of a standard ResNet-50 encoder \cite{he2016deep}, before the final mean-pooling layer, such that hiddens form a 7$\times$7 grid of 2048-dimensional vectors $\vh[i,j]$. For every mask $\vm$ associated with the image, we compute a mask-pooled hidden vector 
\[ \vh_\vm = \frac{1}{\sum_{i,j} m_{i,j}} \sum_{i,j} m_{i,j} \ \vh[i,j], \]
having spatially downsampled the binary mask to a 7\x7 grid with average pooling. We then transform each of these vectors with a two-layer MLP, yielding non-linear projections $\vz_\vm = g( \vh_\vm ) \in \sR^d$. %Note that we can recover the architecture of SimCLR and BYOL by using a single global mask in this step.

For DetCon$_{S}$ we process both views with the same encoder $f_\theta$ and projection network $g_\theta$, where $\theta$ are the learned parameters. For DetCon$_{B}$ one view is processed with $f_\theta$ and $g_\theta$ and the other with $f_\xi$ and $g_\xi$, where $\xi$ is an exponential moving average of $\theta$. The first view is further transformed with a prediction network $q_\theta$. Here again we reuse the details of SimCLR and BYOL for DetCon$_{S}$ and DetCon$_{B}$ respectively in the definition of the projection and prediction networks (see appendix \ref{sec:app-architecture}). In summary, we represent each view and mask as latents $\vv_\vm$ and $\vv'_{\vm'}$ where
\[ \vv_\vm = g_\theta(\vh_{\vm}), \quad \vv'_{\vm'} = g_\theta(\vh'_{\vm'}) \]
for DetCon$_{S}$ and 
\[ \vv_\vm = q_\theta \circ g_\theta(\vh_{\vm}), \quad \vv'_{\vm'} =g_\xi(\vh'_{\vm'})  \]
for DetCon$_{B}$. We rescale all latents with a temperature hyperparameter $\tau$, such that their norm is equal to $1 / \sqrt{\tau}$, with $\tau = 0.1$. Note that for downstream tasks, we retain only the feature extractor $f_\theta$ and discard all other parts of the network (the prediction and projection heads, as well as any exponential moving averages). 

\vspace{0.5em} \noindent \textbf{Objective: contrastive detection.} Let $\vv_\vm, \vv'_{\vm'}$ be the latents representing masks $\vm, \vm'$ in the views $\vx, \vx'$. The contrastive loss function
\begin{equation}
\label{eq:con}
\ell_{\vm, \vm'} = - \log \frac{\exp( \vv_\vm {\cdot} \vv'_{\vm'})}{\exp(\vv_\vm {\cdot} \vv'_{\vm'})+\sum_\vn \exp(\vv_\vm {\cdot} \vv_{\vn})}
\end{equation}
 defines a prediction task: having observed the projection $\vv_\vm$, learn to recognize the latent $\vv'_{\vm'}$ in the presence of negative samples $\{ \vv_\vn \}$. We include negative samples from different masks in the image and different images in the batch. Note that we make no assumptions regarding these masks, allowing negative masks to overlap with the positive one. 

A natural extension of this loss would be to jointly sample paired masks $\vm, \vm'$ which correspond to the same region in the original image, and maximize the similarity of features representing them
\begin{equation}
\label{eq:detcon1}
\mathcal{L} = \E_{(\vm, \vm') \sim \mathcal{M}} \ \ell_{\vm, \vm'}.
\end{equation}
We make a few practical changes to this objective. First, in order to facilitate batched computation we randomly sample at each iteration a set of 16 (possibly redundant) masks from the variable-sized sets of masks $\{\vm\}$ and $\{\vm'\}$. Second, we densely evaluate the similarity between all pairs of masks and all images, such that each image contributes 16 negative samples to the set $\{ \vv_{\vn} \}$ in equation~\eqref{eq:con}, rather than a single one. We aim to makes these negatives as diverse as possible by choosing masks that roughly match different objects in the scene (Section \ref{sec:method_masks}). Finally, we mask out the loss to only maximize the similarity of paired locations, allowing us to handle cases where a mask is present in one view but not another (see Figure \ref{fig:methods}). Together, these simple modifications bring us to the DetCon objective:
\begin{equation}
\label{eq:detcon2}
\mathcal{L} = \sum_{\vm} \sum_{\vm'} \mathds{1}_{\vm, \vm'} \ell_{\vm, \vm'}
\end{equation}
where the binary variable $\mathds{1}_{\vm, \vm'}$ indicates whether the masks $\vm, \vm'$ correspond to the same underlying region. 

\vspace{0.5em} \noindent \textbf{Optimization.} When pretraining on ImageNet we adopt the optimization details of SimCLR and BYOL for training \dcs and \dcb respectively. When pretraining on COCO we make minor changes to the learning schedule to alleviate overfitting (see appendix \ref{sec:app-optimization}). 

\vspace{0.5em} \noindent \textbf{Computational cost.} The computational requirements of self-supervised learning are largely due to forward and backward passes through the convolutional backbone. For the typical ResNet-50 architecture applied to 224\x224-resolution images, a single forward pass requires approximately 4B FLOPS. The additional projection head in SimCLR and \dcs requires an additional 4M FLOPS. Since we forward 16 hidden vectors through the projection head instead of 1, we increase the computational cost of the forward pass by 67M FLOPS, less than 2\% of total. Together with the added complexity of the contrastive loss, this increase is 5.3\% for \dcs and 11.6\% for DetCon$_{B}$ (see appendix \ref{sec:app-architecture}). Finally, the cost of computing image segmentations is negligible because they can be computed once and reused throughout training. Therefore the increase in complexity of our method relative to the baseline is sufficiently small for us to interchangeably refer to ``training iterations'' and ``computational cost''.

\subsection{Unsupervised mask generation}
\label{sec:method_masks}
\noindent To produce masks required by the DetCon objective, we investigate several segmentation procedures, from simple spatial heuristics to graph-based algorithms from the literature. 

\vspace{0.5em} \noindent \textbf{Spatial heuristic.} The simplest segmentation we consider groups locations based on their spatial proximity only. Specifically, we divide the image into an $n \times n$ grid of non-overlapping, square sub-regions (Figure \ref{fig:masks}, \nth{2} column). %As noted in Section \ref{sec:method_framework}, w
Note that when using a single, global mask ($n=1$), the \dcs objective reverts to SimCLR.

\vspace{0.5em} \noindent \textbf{Image-computable masks: FH.} We also consider the Felzenszwalb-Huttenlocher algorithm \cite{felzenszwalb2004efficient}, a classic segmentation procedure which iteratively merges regions using pixel-based affinity (Figure~\ref{fig:masks}, \nth{3} column). We generate a diverse set of masks by varying two hyperparameters, the scale $s$ and minimum cluster size $c$, using $s \in \{500, 1000, 1500\}$ and $c = s$ when training on COCO and $s = 1000$ when training on ImageNet.

\begin{figure*}
  \begin{minipage}[c]{\textwidth}
    \includegraphics[width=\textwidth]{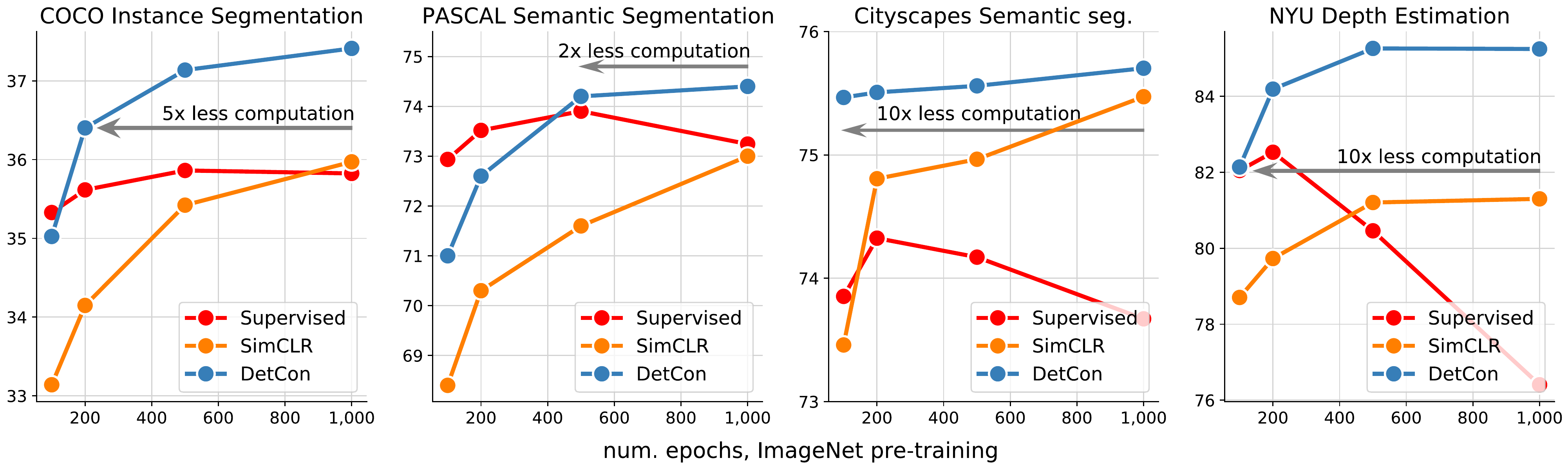}
  \end{minipage} % \hfill
  \begin{minipage}[c]{\textwidth} \vspace{0.5em}
  \caption{\textbf{Efficient ImageNet pretraining with DetCon$_{S}$.} We pretrain networks with SimCLR, DetCon$_{S}$, or supervised learning on ImageNet for different numbers of epochs, and fine-tune them for COCO detection and instance segmentation (for 12 epochs), semantic segmentation on PASCAL or Cityscapes, or depth estimation on NYU v2. \dcs outperforms SimCLR, with up to 10\x \ less pretraining.} \label{fig:transfer-from-imagenet}
  \end{minipage}
\end{figure*}

\vspace{0.5em} \noindent \textbf{Image-computable masks: MCG.} Multiscale Combinatorial Grouping \cite{arbelaez2014multiscale} is a more sophisticated algorithm which groups superpixels into many overlapping object proposal regions~\cite{carreira2011cpmc}, guided by mid-level  classifiers (Figure~\ref{fig:masks}, \nth{4} column). For each image we use 16 MCG masks with the highest scores. Note that the fact that these masks can overlap is supported by our formulation. 

\vspace{0.5em} \noindent \textbf{Human annotated masks.} Throughout this work we consider the benefits afforded by the use of the \textit{unsupervised} masks detailed above. In the final section, we ask whether higher quality masks (provided by human annotators; Figure \ref{fig:masks}, \nth{5} column) can improve our results. 

\begin{table*}[t]
\small
\begin{tabularx}{\textwidth}{c *{10}{Y}}

 & \multicolumn{2}{c}{Detection}  
 & \multicolumn{2}{c}{Instance Segmentation}
 & \multicolumn{4}{c}{Semantic Segmentation}
 & \multicolumn{2}{c}{Depth Estimation}
 \\

 & \multicolumn{2}{c}{COCO}  
 & \multicolumn{2}{c}{COCO}
 & \multicolumn{2}{c}{PASCAL}
 & \multicolumn{2}{c}{Cityscapes}
 & \multicolumn{2}{c}{NYU v2}
 \\
\cmidrule(lr){2-3} \cmidrule(l){4-5} \cmidrule(l){6-7} \cmidrule(l){8-9} \cmidrule(l){10-11}
Pretrain epochs & 300 & 1000 & 300 & 1000 & 300 & 1000  & 300 & 1000  & 100 & 1000 \\
\midrule
 BYOL &
41.2 & \textcolor{brown}{41.6} & 
37.1 & \textcolor{purple}{37.2} & 
74.7 & \textcolor{blue}{75.7} & 
73.4 & \textcolor{violet}{74.6} & 
83.7 & \textcolor{teal}{84.2} \\
\textbf{DetCon$_{B}$} &
\textcolor{brown}{42.0} & 42.7 & 
\textcolor{purple}{37.8} & 38.2 & 
\textcolor{blue}{75.6} & 77.3 & 
\textcolor{violet}{75.1} & 77.0 & 
\textcolor{teal}{85.1} & 86.3 \\
Efficiency Gain &
\multicolumn{2}{c}{\textcolor{brown}{$>3\times$}} & \multicolumn{2}{c}{\textcolor{purple}{$>3\times$}} & \multicolumn{2}{c}{\textcolor{blue}{$\approx3\times$}} &
\multicolumn{2}{c}{\textcolor{violet}{$>3\times$}} & 
\multicolumn{2}{c}{\textcolor{teal}{$>10\times$}} \\ 
\end{tabularx}

\vspace{.5em}
\caption{\textbf{Efficient ImageNet pretraining with DetCon$_{B}$.} We pretrain networks on ImageNet with BYOL or DetCon$_{B}$, and fine-tune them for COCO detection and instance segmentation (for 12 epochs), semantic segmentation on PASCAL or Cityscapes, or depth estimation on NYU v2. \dcb outperforms BYOL, with up to 10\x \ less pretraining (colors highlight gains in pretraining efficiency). }
\label{tab:comp_simclr_byol_imagenet}
\vspace{-1.em}
\end{table*}

\subsection{Evaluation protocol}
\label{sec:method_eval}
\noindent Having trained a feature extractor in an unsupervised manner, we evaluate the quality of the representation by fine-tuning it for object detection and instance segmentation on COCO, segmentatic segmentation  on PASCAL and Cityscapes, and depth estimation on NYU v2.

\vspace{0.5em} \noindent \textbf{Object detection and instance segmentation.} We use the pretrained network to initialize the feature extractor of a Mask-RCNN \cite{he2017mask} equipped with feature pyramid networks \cite{lin2017feature} and cross-replica batch-norm \cite{peng2018megdet}. We adopt the Cloud TPU implementation\footnote{\url{https://github.com/tensorflow/tpu/tree/master/models/official/detection}} and use it without modification. We fine-tune the entire model on the COCO \texttt{train2017} set, and report bounding-box AP (\apbbox{}) and mask AP (\apmask{}) on the \texttt{val2017} set. We use two standard training schedules: 12 epochs and 24 epochs \cite{he2019momentum}.

\vspace{0.5em} \noindent \textbf{Semantic segmentation.} Following \cite{he2019momentum} we use our network to initialize the backbone of a fully-convolutional network \cite{long2015fully}. For PASCAL, we fine-tune on the \texttt{train\_aug2012} set for 45 epochs and report the mean intersection over union (mIoU) on the \texttt{val2012} set. For Cityscapes, we finetune on the \texttt{train\_fine} set for 160 epochs and evaluate on the  \texttt{val\_fine} set.

\vspace{0.5em} \noindent \textbf{Depth estimation.} Following \cite{grill2020bootstrap} we stack the deconvolutional network from \cite{laina2016deeper} on top of our feature extractor, and fine-tune on the NYU v2 dataset. We report accuracy as the percentage of errors below 1.25 (pct$<$1.25).

\section{Experiments}

\noindent Our main self-supervised learning experiments employ FH masks, because as we will show, with DetCon they outperform simple spatial heuristics and approach the performance of MCG masks while being fast and easy to apply to large datasets such as ImageNet, given their availability in  \texttt{scikit-image} \cite{scikit-image}.

\subsection{Transfer learning from ImageNet}
\label{sec:exp_imagenet}

\noindent We first study whether the DetCon objective improves the pretraining efficiency of transfer learning from ImageNet. 

\vspace{0.5em} \noindent \textbf{Pretraining efficiency.}  We train SimCLR and DetCon$_{S}$ models on ImageNet for 100, 200, 500 and 1000 epochs, and transfer them to several datasets and tasks. Across all downstream tasks and pretraining regimes, \dcs substantially outperforms SimCLR (Figures \ref{fig:intro} and \ref{fig:transfer-from-imagenet}, blue and orange curves). When fine-tuning on COCO, %DetCon$_{S}$ substantially outperforms SimCLR across all training regimes (Figure \ref{fig:intro}; Figure \ref{fig:transfer-from-imagenet}, \nth{1} column). Equivalently, 
the performance afforded by 1,000 epochs of SimCLR pretraining is surpassed by only 200 epochs of \dcs pretraining (i.e.\ a 5\x \ gain in pretraining efficiency). We found similar results when transferring to other downstream tasks:  DetCon$_{S}$ yields a 2\x \ gain in pretraining efficiency for PASCAL semantic segmentation (Figure \ref{fig:transfer-from-imagenet}, \nth{2} column), and 10\x \ gains for Cityscapes semantic segmentation and NYU depth prediction. (Figure \ref{fig:transfer-from-imagenet}, \nth{3}, and \nth{4} columns). 

We also evaluated the transfer performance of a supervised ResNet-50 trained on ImageNet (Figures \ref{fig:intro} and \ref{fig:transfer-from-imagenet}, red curve). While supervised pretraining performs well with small computational budgets (e.g.\ 100 pretraining epochs), it quickly saturates, indicating that ImageNet labels only partially inform downstream tasks. This is emphasized for Cityscapes semantic segmentation and NYU depth prediction, which represent a larger shift in the domain and task.

\vspace{0.5em} \noindent \textbf{From BYOL to DetCon$_{B}$.} How general is DetCon? We tested this by comparing \dcb to the BYOL framework, upon which it is based. We adopt the underlying framework details (e.g.\ data-augmentation, architecture, and optimization) \textit{without modification}, possibly putting the DetCon objective at a disadvantage. Despite this, DetCon$_{B}$ outperforms BYOL across pretaining budgets and downstream tasks. In particular, \dcb yields a 3\x \ gain in pretraining efficiency when transferring to COCO, PASCAL, and Cityscapes detection and segmentation, and a 10\x \ gain when transferring to NYU depth prediction (Table \ref{tab:comp_simclr_byol_imagenet}).

\vspace{0.5em} \noindent \textbf{Comparison with prior art.} We now compare to other works in self-supervised transfer learning, and use fully-trained DetCon$_{S}$ and DetCon$_{B}$ models for the comparison. Here we focus on transfer to COCO as it is more widely studied. Note that other methods use a slightly different implementation of the Mask-RCNN \cite{wu2019detectron2} however their results for supervised ImageNet pretraining and SimCLR match our own \cite{tian2020makes, xie2020propagate}, enabling a fair comparison. Table \ref{tab:prior_art_r50} shows that DetCon outperforms all other methods for supervised and self-supervised transfer learning. 

\begin{table}[t]
\small
\begin{tabularx}{\linewidth}{c *{4}{Y}}
 & \multicolumn{2}{c}{Fine-tune 1\x}  
 & \multicolumn{2}{c}{Fine-tune 2\x}
 \\
\cmidrule(lr){2-3} \cmidrule(l){4-5} 
method & \apbbox{~} & \apmask{~} & \apbbox{~} & \apmask{~} \\
\midrule
Supervised & 39.6 & 35.6 & 41.6 & 37.6 \\
VADeR \cite{pinheiro2020unsupervised}  & 39.2 & 35.6 & - & - \\
MoCo \cite{he2019momentum}  & 39.4 & 35.6 & 41.7 & 37.5 \\
SimCLR \cite{chen2020simple}  & 39.7 & 35.8 & 41.6 & 37.4 \\ 
MoCo v2 \cite{chen2020improved}  & 40.1 & 36.3 & 41.7 & 37.6 \\ 
InfoMin \cite{tian2020makes}  & 40.6 & 36.7 & 42.5 & 38.4 \\ 
PixPro \cite{xie2020propagate}  & 41.4 & - &  - &  - \\ 
BYOL \cite{grill2020bootstrap}  & 41.6 & 37.2 & 42.4 & 38.0 \\ 
SwAV \cite{caron2020unsupervised}  & 41.6 & 37.8 & - & - \\ 
% \hline
\midrule
\textbf{DetCon$_{S}$} & 41.8 & 37.4 & 42.9 & 38.1  \\ 
\textbf{DetCon$_{B}$} & \textbf{42.7} & \textbf{38.2} & \textbf{43.4} & \textbf{38.7} \\ 
\end{tabularx}

\vspace{.5em}
\caption{\textbf{Comparison to prior art:} all methods are pretrained on ImageNet then fined-tuned on COCO for 12 epochs (1\x ~ schedule) or 24 epochs (2\x ~ schedule). %Bounding-box AP (\apbbox{}) and mask AP (\apmask{}) are evaluated on the COCO \texttt{val2017} set.
}
\label{tab:prior_art_r50}
\vspace{-1.em}
\end{table}

\begin{figure}[t]
    \begin{center}
        \includegraphics[width=\linewidth]{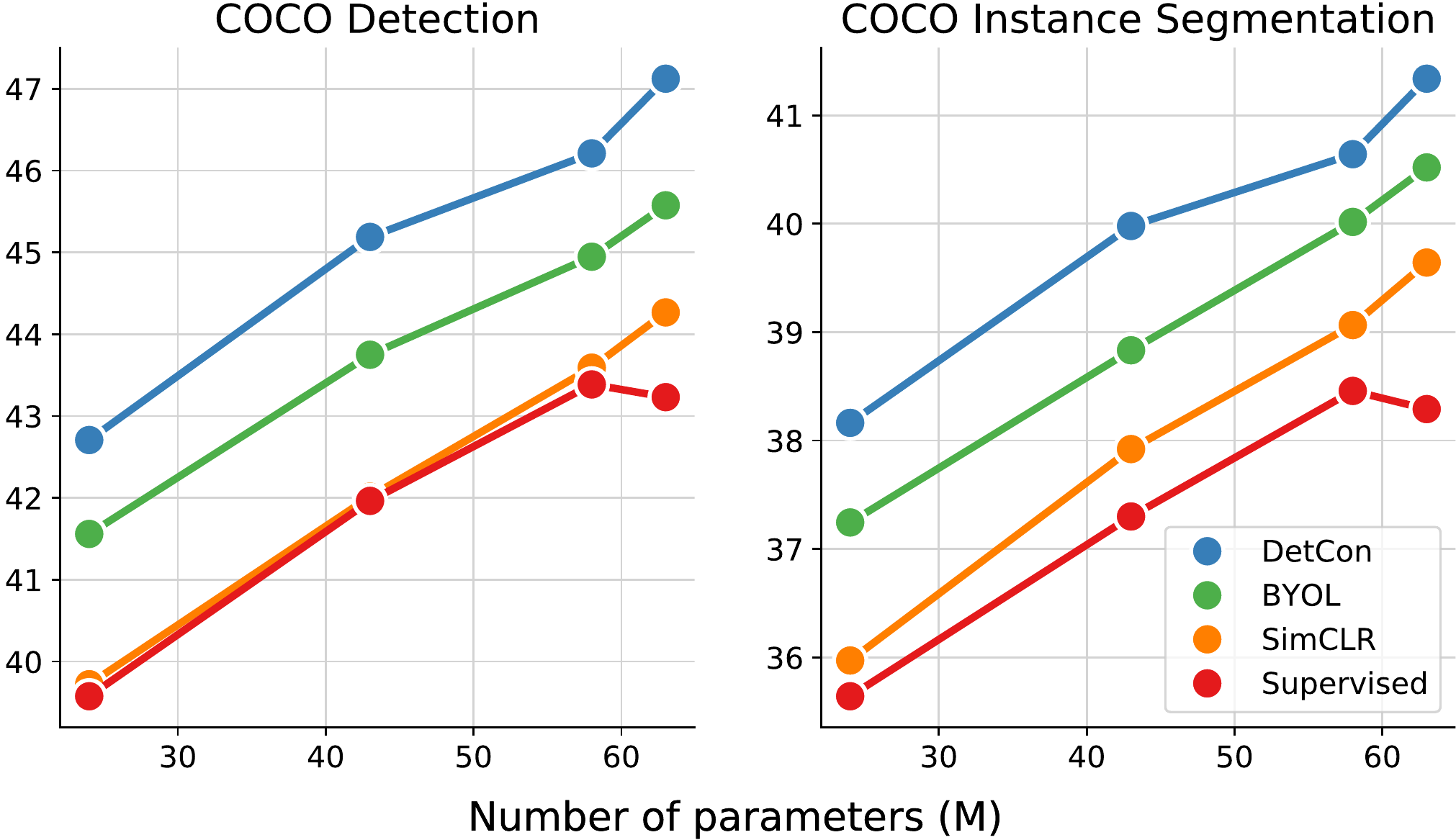}
    \end{center}
  \caption{\textbf{Scaling DetCon to larger models.} We pretrain ResNet-50, ResNet-101, ResNet-152, and ResNet-200 feature extractors on ImageNet using supervised learning, SimCLR, BYOL, or \dcb and fine-tune them on COCO for 12 epochs.
  } \label{fig:big_models}
  \vspace{-1em}
\end{figure}

\vspace{0.5em} \noindent \textbf{Scaling model capacity.} Prior works in self-supervised learning have been shown to scale very well with model capacity \cite{doersch2017multi, kolesnikov2019revisiting, chen2020simple}. Could the gains afforded by DetCon disappear with larger models? We trained SimCLR, BYOL, and DetCon$_{B}$ models on ImageNet, using ResNet-101, -152, and -200 feature extractors instead of ResNet-50. %Tables \ref{tab:prior_art_r101} and \ref{tab:prior_art_r200}, and 
Figure \ref{fig:big_models} and Table \ref{tab:prior_art_big} show that DetCon continues to outperform other methods in this higher-capacity regime.

We went a step further and trained a ResNet-200 with a 2\x \ width multiplier \cite{kolesnikov2019revisiting}, containing 250M parameters. Surprisingly, despite only being trained on ImageNet, this model's transfer performance matches that of a very recently proposed large-scale self-supervised model, SEER~\cite{goyal2021self}, having 693M parameters and trained on 1000\x \ more data (Table \ref{tab:big_boys}). While the comparison is imperfect (large-scale data is necessarily more noisy), it highlights the potential of improvements from the self-supervised learning objective alone. 

\begin{table}[h]
\small
\vspace{-0.5em}
\begin{tabularx}{\linewidth}{c *{4}{Y}}
pretrain & Data & Params & \apbbox{~} & \apmask{~} \\
\midrule
Supervised \cite{goyal2021self}& IN-1M  & 250 M & 45.9 & 41.0 \\
~ ~ SEER \cite{goyal2021self}&  \textbf{IG-1B}\hspace{0.125em} & \textbf{693 M} & 48.5 & \textbf{43.2} \\
\midrule
\textbf{DetCon$_{B}$} &  IN-1M  & 250 M & \textbf{48.9} & 43.0 \\ 
\end{tabularx}

\vspace{.5em}
\caption{\textbf{Comparison to large-scale transfer learning:} all methods pretrain a backbone and transfer to COCO detection and instance segmentation using a Mask-RCNN. SEER trains on a billion Instagram images whereas \dcs trains on ImageNet (1.3 million images). SEER and the supervised baseline use the recent RegNet architecture \cite{radosavovic2020designing}, whereas \dcs uses a generic ResNet-200 (2\x \ width). Despite this, DetCon pretraining matches the performance of large-scale SEER pretraining.
 } \label{tab:big_boys}
\vspace{-0.5em}
\end{table}

\subsection{Transfer learning from COCO}
\label{sec:exp_coco}

\noindent We next investigate the ability of the DetCon objective to handle complex scenes with multiple objects. For this we pretrain on the COCO dataset and compare to SimCLR. 

\begin{figure}[h]
  \begin{minipage}[c]{0.58\linewidth}
    \includegraphics[width=\linewidth]{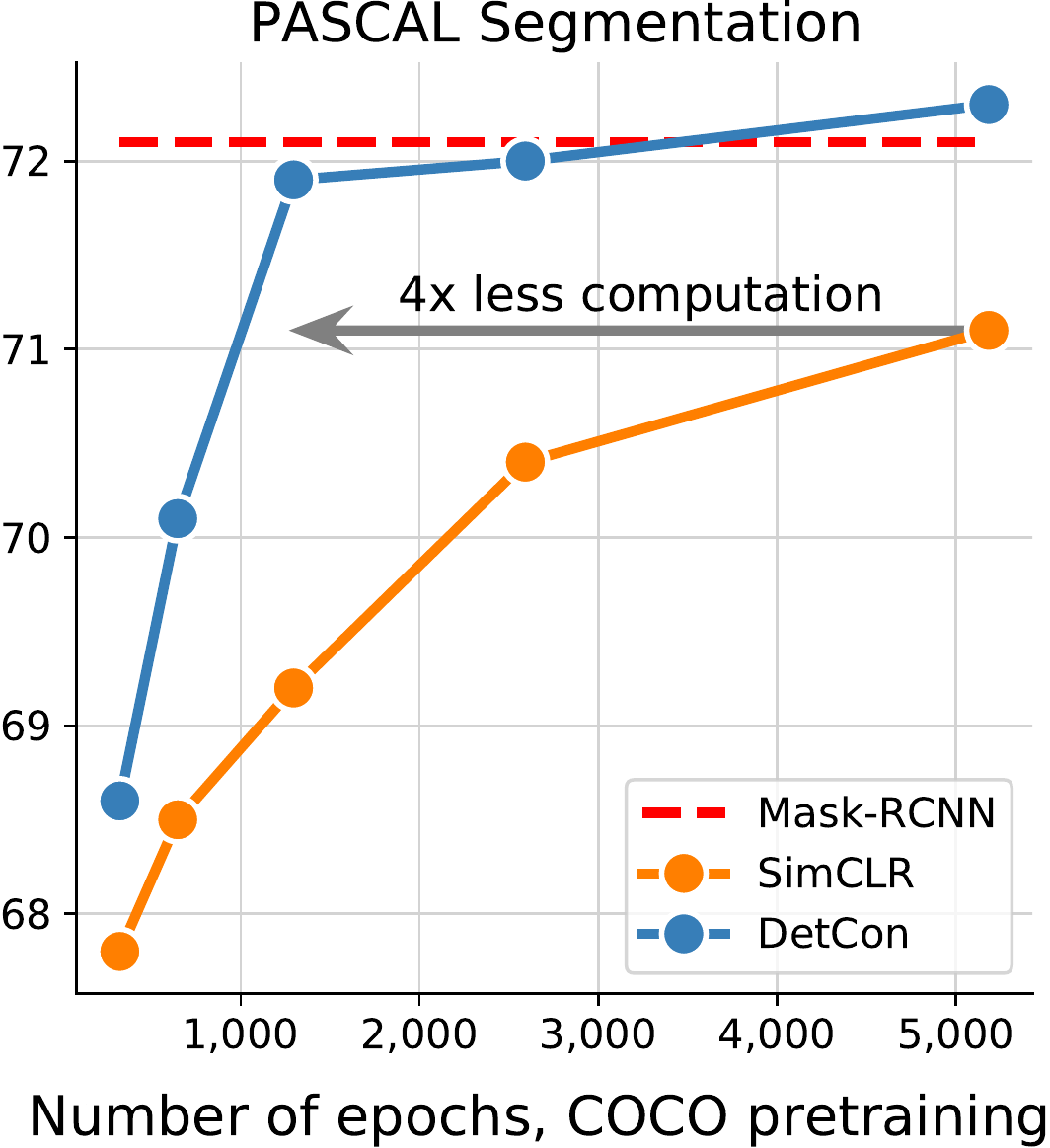}
  \end{minipage}\hfill
  \begin{minipage}[c]{0.4\linewidth}
  \caption{\textbf{Efficient transfer from COCO.} We pretrain representations using SimCLR or DetCon$_{S}$ on COCO for different numbers of epochs, and transfer to PASCAL semantic segmentation by fine-tuning them for 45 epochs.\vspace{6em}} \label{fig:transfer-from-coco}
  \end{minipage}
\vspace{1em}
\end{figure}

\vspace{0.5em} \noindent \textbf{Training efficiency.} We train SimCLR and DetCon$_{S}$ for a range of schedules (324--5184 epochs), and transfer all models to semantic segmentation on PASCAL. We find DetCon$_{S}$ to outperform SimCLR across training budgets (Figure \ref{fig:transfer-from-coco}). As before, the maximum accuracy attained by SimCLR is reached with 4\x \ less pretraining time.

\vspace{0.5em} \noindent \textbf{Surpassing supervised transfer from COCO.} We also evaluated the transfer performance of representations trained on COCO in a supervised manner. Specifically, we trained a Mask-RCNN with a long schedule (108 epochs, a ``9\x'' schedule), and use the learned feature extractor (a ResNet-50, as for SimCLR and DetCon pretraining) as a representation for PASCAL segmentation. Unlike SimCLR, DetCon pretraining surpasses the performance of this fully supervised baseline (Figure \ref{fig:transfer-from-coco}).  % , SimCLR continues to lag behind

\begin{figure}[t]
    \vspace{-2em}
    \begin{center}
        \includegraphics[width=0.9\linewidth]{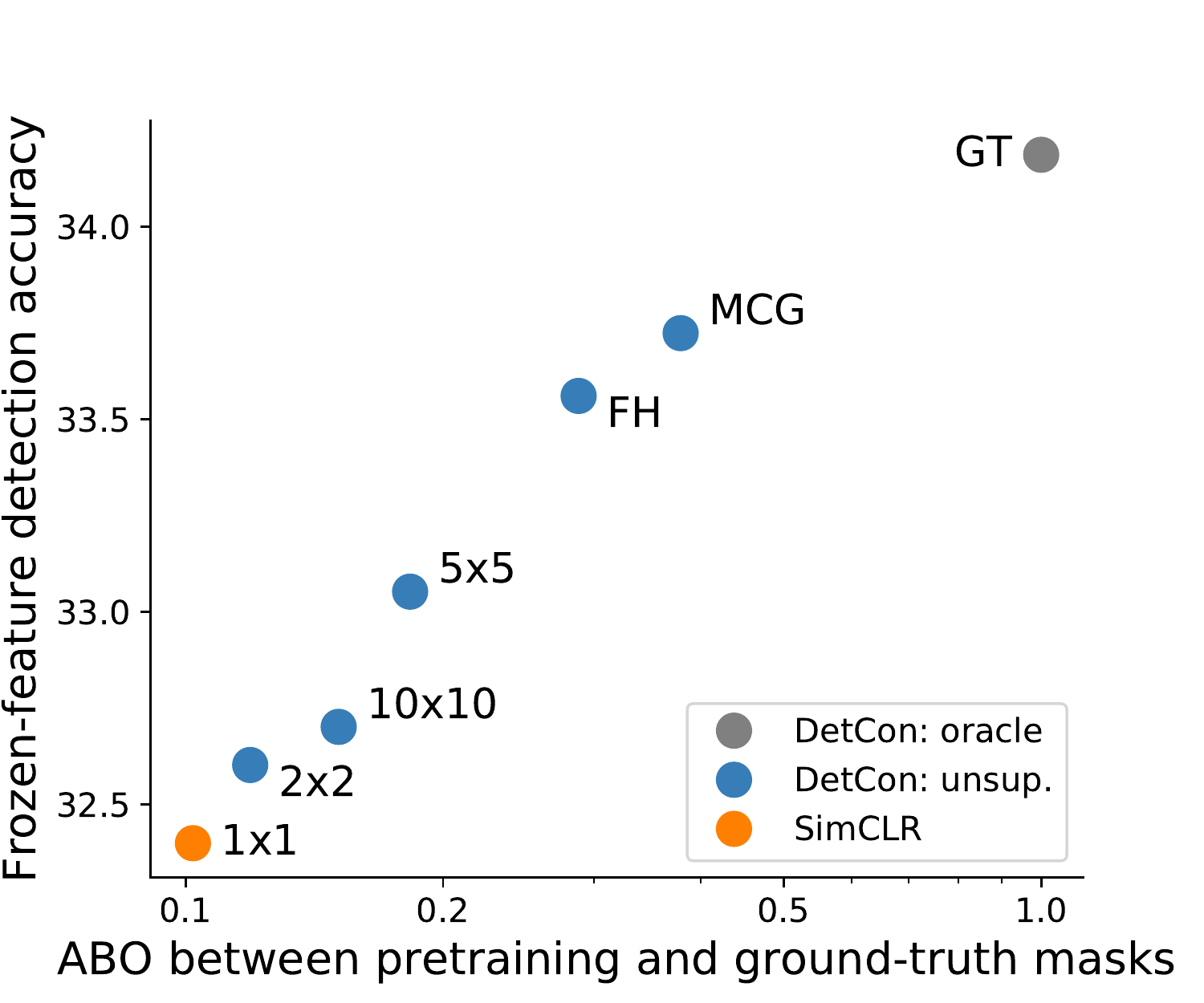}
    \end{center}
    \vspace{-1em}
   \caption{\textbf{Effect of type of masks used in DetCon objective.} We train DetCon models on COCO using unsupervised masks (blue), or the ground-truth COCO masks (grey). Using a single, global mask (i.e.\ a ``1\x1'' grid) is equivalent to SimCLR (orange). We compute the Average Best Overlap (ABO) by measuring the IoU between each ground-truth mask and the closest pretraining mask, and averaging over all ground truth instances and images (x-axis). We  evaluate the accuracy of each model on COCO detection using the \textit{frozen-feature} paradigm (y-axis).}
\label{fig:ablation}
\vspace{-1em}
\end{figure}

\vspace{-1em}
\subsection{Ablations and analysis}

\noindent We now dissect the components of the DetCon objective and assess the benefits of each. For this we pretrain on COCO as it contains complex scenes with many objects and associated ground-truth masks, allowing us to measure the impact of segmenting them accurately. We evaluate learned representations with a \textit{frozen-feature analysis}, in which the feature extractor is kept fixed while we train the other layers of a Mask-RCNN also on COCO. This controlled setting is analogous\footnote{But note that the Mask-RCNN contains several non-linear layers due to the additional complexity of the output space relative to classification.} to the linear classification protocol used to evaluate the quality of self-supervised representations for image recognition \cite{doersch2015unsupervised, donahue2016adversarial, pathak2016learning, zhang2016colorful}

\label{sec:exp_ablations}

\vspace{0.5em} \noindent \textbf{What makes good masks?} The DetCon objective can be used with a variety of different image segmentations, which ones lead to the best representation? We first consider spatial heuristics which partition the image into 2\x2, 5\x5, or 10\x10 grids, a 1\x1 grid being equivalent to using the SimCLR objective. We find downstream performance increases with finer grids, a 5\x5 grid being optimal (Figure \ref{fig:ablation}). 

Next we consider image-computable FH and MCG masks, both of which outperform the spatial heuristic masks, MCG masks leading to slightly better representations. Interestingly, the quality of the representation \textit{correlates very well with the overlap between pretraining masks and ground-truth}---the better each ground truth object is covered by some mask, the better DetCon performs.

\begin{table}[b]
\small
\vspace{-1.0em}
\begin{tabularx}{\linewidth}{c *{4}{Y}}
model & masks & \#latents & \apbbox{~} & \apmask{~} \\
\midrule

SimCLR                & global & ~ 1 & 31.6 & 29.2 \\
(a)                   & global &  16 & 31.5 & 29.3 \\
(b)                   & FH     & ~ 1 & 31.2 & 28.8 \\
\midrule
\textbf{DetCon}$_{S}$ & FH     &  16 & \textbf{33.4} & \textbf{30.6} \\
\end{tabularx}

\vspace{.5em}
\caption{\textbf{Ablation: from SimCLR to DetCon$_{S}$}. We pretrain on COCO and evaluate \textit{frozen feature} accuracy also on COCO. \textbf{masks:} specifies whether hidden vectors are pooled globally, or within individual FH masks. \textbf{\#latents:} number of masks.}
\label{tab:ablation}
\end{table}

\vspace{0.5em} \noindent \textbf{Contrastive detection vs contrastive recognition.} 
How does the DetCon objective benefit from these image segmentations? We assess the impact of each of its components by incrementally adding them to the SimCLR framework. As mentioned previously, we recover SimCLR when using a single, global mask in the DetCon objective. As a sanity check, we verify that duplicating this mask several times and including the resulting (identical) features in the DetCon objective makes no difference in the quality of the representation (Table \ref{tab:ablation}, row \textbf{a}). Interestingly, using FH masks but only sampling a single mask per image slightly deteriorates performance, presumably because the model only learns from part of the image at every iteration (Table \ref{tab:ablation}, row \textbf{b}). By densely sampling object regions \dcs learns from the entire image, while also benefiting from a diverse set of positive and negative samples, resulting in increased detection and segmentation accuracy (Table \ref{tab:ablation}, final row).

\subsection{What if segmentation were solved?}
\label{sec:gt}
\noindent The DetCon objective function leads to fast transfer learning and strong performance despite using fairly approximate segmentation masks. In Section \ref{sec:exp_ablations} we found higher quality segmentations (such as those computed using MCG, or obtained from human annotators) to improve representational quality. How might we improve the learning objective given more accurate segmentations? We assessed this question by revisiting our design choices for the contrastive objective, when given ground-truth masks from the COCO dataset as opposed to the approximate FH masks.

\begin{figure}[t]
  \begin{minipage}[c]{0.58\linewidth}
  \includegraphics[width=\linewidth]{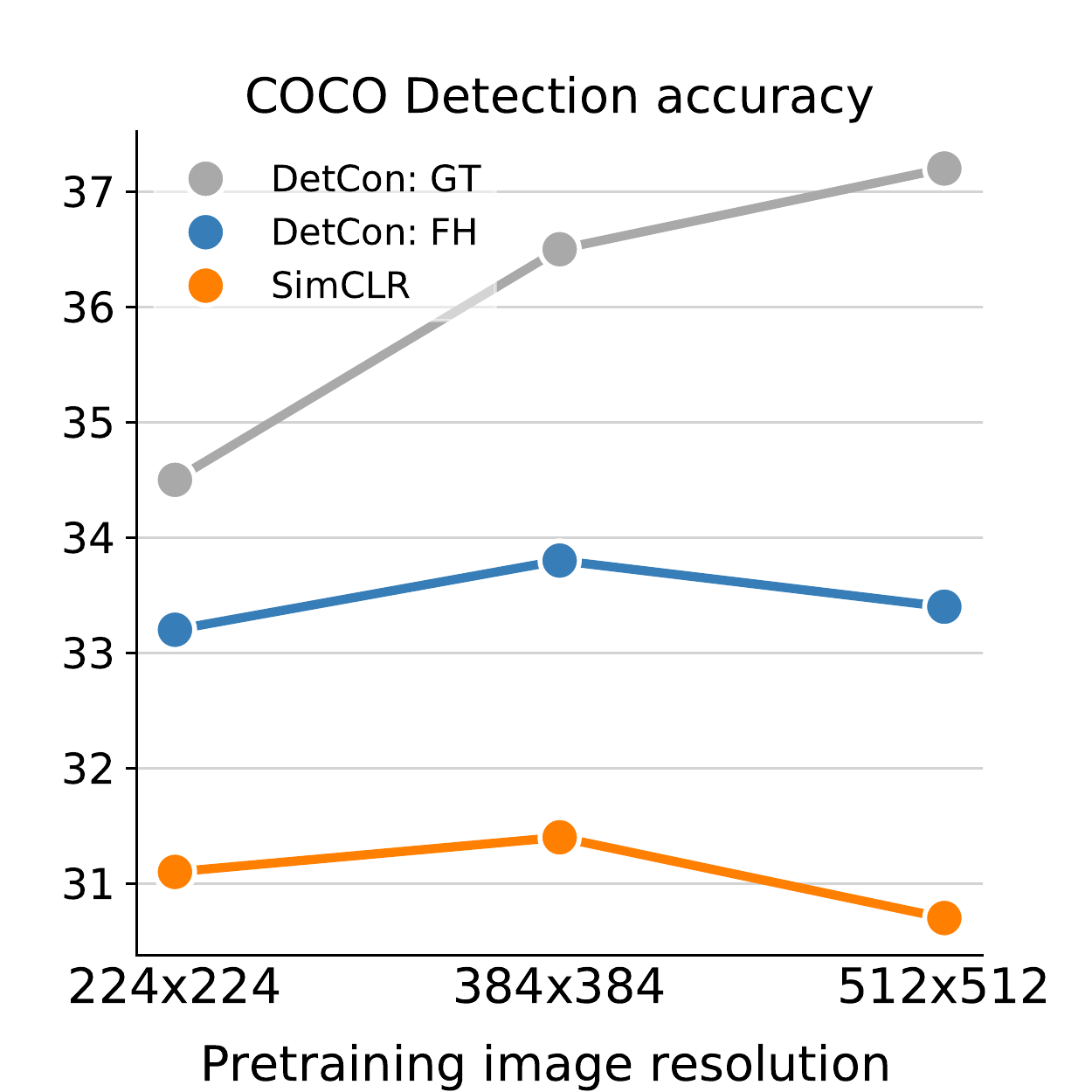}
  \end{minipage}\hfill
\begin{minipage}[c]{0.4\linewidth}
  \caption{
  \textbf{Better segmentations benefit from higher resolutions.}
  We pretrain backbone networks on COCO using SimCLR and DetCon$_{S}$ (with FH or GT masks) at various resolutions. We report \textit{frozen-feature} performance with a fixed resolution of 1024\x1024.} \label{fig:scaling}
  \end{minipage}
\vspace{-4mm}
\end{figure}

\vspace{0.5em} \noindent \textbf{Scaling image resolution.} We hypothesized that higher image resolutions might enable the network to benefit more from these more informative segmentations. To preserve fine-grained information we sample local features from within each mask and optimize them using the DetCon objective. We pretrain SimCLR and DetCon$_S$ models equipped with FH or ground-truth (GT) masks, given 384\x384 or 512\x512 resolution images%, keeping all other parameters the same
. While DetCon with FH masks only modestly benefited and SimCLR's performance deteriorated with high-resolution images, DetCon with GT masks improves substantially (Figure \ref{fig:scaling}). Note that this is solely due to an improved representation quality; the image resolution used for downstream evaluation is maintained at 1024\x1024 for all models. 

\vspace{0.5em} \noindent \textbf{Revisiting the contrastive framework.} Finally, we asked whether the current contrastive learning paradigm---which utilizes large numbers of negatives and predictions across stochastic augmentations---remains optimal in the context of the DetCon objective with high-quality segmentations.

\vspace{0.25em} \textbf{\textit{Are large numbers of negative samples necessary?}} Not with high-quality masks. When dividing the total number of negative samples by 128 (by only gathering negatives from within a worker) the performance of DetCon with FH masks drops (Table \ref{tab:gt_only}, row \textbf{a}), consistently with other contrastive learning frameworks \cite{chen2020simple, he2019momentum}. In contrast, DetCon$_{S}$ using GT masks improves despite this limitation. 

\vspace{0.25em} \textbf{\textit{Are positive pairs sampled across augmented views necessary?}} Not with high-quality masks. We run DetCon models while sampling a single augmentation for each image and maximizing the similarity of mask-based features \textit{within} this view. Here again, the DetCon objective suffers from this handicap when using approximate FH masks, but not with high-quality segmentations (Table \ref{tab:gt_only}, row \textbf{b}).

\vspace{0.25em} \textbf{\textit{How can this be?}} One interpretation is that other images give us clean negative examples because images in COCO depict different scenes. %, but this does not mean these are higher quality than negative examples from the same image.
However it appears that negatives from the same image provide a stronger learning signal (in that they share features such as lighting, background, etc) as long as they are clean---i.e., we are not pushing features from the same object apart. Positives from the same image are also at least as good as those across augmentations if again they are clean---i.e., we are not pulling together features from different objects.

\begin{table}[t]
\small
% \vspace{-1.0em}
\begin{tabularx}{\linewidth}{c *{4}{Y}}

 &  &  & \multicolumn{2}{c}{Masks} \\
 \cmidrule(lr){4-5}
model & all neg & two views & FH & GT \\
\midrule

DetCon & \checkmark & \checkmark & \textbf{33.6} & 37.0 \\
(a)    &            & \checkmark & 32.2 & 38.5 \\
(b)    &            &            & 27.7 & \textbf{38.8} \\
\end{tabularx}

\vspace{.5em}
\caption{\textbf{Simplifying the contrastive framework.} We train \dcs models on COCO using approximate FH masks or higher-quality ground-truth (GT) masks, and evaluate them in the \textit{frozen-feature} setting. \textbf{``all neg'':} Negative samples are collected from the entire batch as opposed to only within a worker (out of 128 workers). \textbf{``two views'':} Contrastive predictions are made across augmentations, as opposed to within a view.}
\vspace{-1.0em}
\label{tab:gt_only}
\end{table}

\section{Discussion}

\noindent We have proposed DetCon, a simple but powerful self-supervised learning algorithm. %variation on existing self-supervised learning algorithms such as SimCLR and BYOL. 
By exploiting low-level cues for organizing images into entities such as objects and background regions, DetCon accelerates %boosts the efficiency of 
pretraining on large datasets %by up to 5\x, 
while %also 
improving %the 
accuracy on a variety of % of the learned representations on 
downstream tasks. Our best models achieve state-of-the-art performance among self-supervised methods pretrained on ImageNet and match % similar performance to a 
recent state-of-the-art methods training larger models on a much larger dataset~\cite{goyal2021self}.

We showed that the power of DetCon strongly correlates with how well the masks used align with object boundaries. This seems intuitive---the DetCon objective can only leverage independent learning signals from each image region if they contain distinct content. Similarly, the resulting negative samples are genuinely diverse only if they represent different objects. This opens exciting prospects of research in jointly discovering objects and learning to represent them. Given the improved performance of DetCon representations for instance segmentation, a natural question is whether they could be used to perform better unsupervised segmentations than the ones used during pretraining. If so, these might be used to learn better representations still, leading to a virtuous crescendo of unsupervised scene understanding.

\section*{Acknowledgements}

\noindent The authors are grateful to Carl Doersch, Raia Hadsell, and Evan Shelhamer for insightful discussions and feedback on the manuscript. 

\appendix
\counterwithin{table}{section}
% \newpage ~
% \newpage 

% \twocolumn[
% \begin{center}
% {\Large \bf \vspace{-0.5em} Supplementary Material: \\ \vspace{0.5em} Efficient Visual Pretraining with Contrastive Detection}
% \end{center}
% ]

\section{Appendix}

\subsection{Implementation: data augmentation}
\label{sec:app-data}

\noindent \textbf{Self-supervised pretraining.} Each image is randomly augmented twice, resulting in two images: $\vx, \vx'$. The augmentations are constructed as compositions of the following operations, each applied with a given probability:
\begin{enumerate}[itemsep=0mm]
\item random cropping: a random patch of the image is selected, whose area is uniformly sampled in $[0.08 \cdot \mathcal{A}, \mathcal{A}]$, where $\mathcal{A}$ is the area of the original image, and whose aspect ratio is logarithmically sampled in $[3/4, 4/3]$. The patch is then resized to 224 \x 224 pixels using bicubic interpolation;
\item horizontal flipping;
\item color jittering: the brightness, contrast, saturation and hue are shifted by a uniformly distributed offset;
\item color dropping: the RGB image is replaced by its grey-scale values;
\item gaussian blurring with a 23\x23 square kernel and a standard deviation uniformly sampled from $[0.1, 2.0]$;
\item solarization: a point-wise color transformation $x \mapsto x \cdot \mathds{1}_{x < 0.5} + (1 - x) \cdot \mathds{1}_{x \ge 0.5}$ with pixels $x$ in $[0, 1]$.
\end{enumerate}
\noindent The augmented images $\vx, \vx'$ result from augmentations sampled from distributions $\mathcal{T}$ and~$\mathcal{T}'$ respectively. These distributions apply the primitives described above with different probabilities, and different magnitudes. The following table specifies these parameters for the SimCLR \cite{chen2020simple} and BYOL frameworks \cite{grill2020bootstrap}, which we adopt for \dcs and \dcb without modification. 
\begin{table}[ht]
    \small
    \centering
    \begin{tabular}{l c c c c }
                                        & \multicolumn{2}{c}{\dcs} & \multicolumn{2}{c}{\dcb} \\
    Parameter                           & $\mathcal{T}$ & $\mathcal{T}'$ & $\mathcal{T}$ & $\mathcal{T}'$ \\ \hline
    Random crop probability             & \multicolumn{4}{c}{1.0} \\
    Flip probability                    & \multicolumn{4}{c}{0.5} \\
    Color jittering probability         & \multicolumn{4}{c}{0.8} \\
    Color dropping probability          & \multicolumn{4}{c}{0.2} \\
    Brightness adjustment max           & \multicolumn{2}{c}{0.8} & \multicolumn{2}{c}{0.4} \\
    Contrast adjustment max             & \multicolumn{2}{c}{0.8} & \multicolumn{2}{c}{0.4} \\
    Saturation adjustment max           & \multicolumn{2}{c}{0.8} & \multicolumn{2}{c}{0.2} \\
    Hue adjustment max                  & \multicolumn{2}{c}{0.2} & \multicolumn{2}{c}{0.1} \\
    Gaussian blurring probability       & $1.0$ & $0.0$ & $1.0$ & $0.1$ \\
    Solarization probability            & $0.0$ & $0.0$ & $0.0$ & $0.2$ \\
    \end{tabular}
     \vspace{0.5em}
\end{table}

\noindent \textbf{Transfer to COCO.} When fine-tuning, image are randomly flipped and resized to a resolution of $u \cdot 1024$ pixels on the longest side, where $u$ is uniformly sampled in $[0.8, 1.25]$, then cropped or padded to a 1024\x1024 image. The aspect ratio is kept the same as the original image. During testing, images are resized to 1024 pixels on the longest side then padded to 1024\x1024 pixels. 

\vspace{0.5em} \noindent \textbf{Transfer to PASCAL.} During training, images are randomly flipped and scaled by a factor in $[0.5, 2.0]$. Training and testing are performed with 513\x513-resolution images.  

\vspace{0.5em} \noindent \textbf{Transfer to Cityscapes.} During training, images are randomly horizontally flipped and scaled by a factor in $[0.5, 2.0]$, with minimum step size 0.25 within that range. Training is performed on 769\x769-resolution images and testing is performed on 1025\x2049-resolution images.

\vspace{0.5em} \noindent \textbf{Transfer to NYU-Depth v2.} The original 640\x480 frames are down-sampled by a factor of 2 and center-cropped to
304\x228 pixels. For training, images are randomly flipped horizontally and color jittered with the same grayscale, brightness, saturation, and hue settings as \cite{grill2020bootstrap}.

\subsection{Implementation: architecture}
\label{sec:app-architecture}

\noindent Our default feature extractor is a ResNet-50 \cite{he2016deep}. In Section \ref{sec:exp_imagenet} we also investigate deeper architectures (ResNet-101, -152, and -200), and a wider model (ResNet-200 \x2) obtained by scaling all channel dimensions by a factor of 2. 

As detailed in Section \ref{sec:method_framework}, this encoder yields a grid of hidden vectors which we pool within masks to obtain a set of vectors $\vh_{\vm}$ representing each mask. These are then transformed by a projection head $g$ (and optionally a prediction head $q$) before entering the contrastive loss. 

%   in \dcs
\vspace{0.5em} \noindent \textbf{DetCon$_{S}$.} Following SimCLR, the projection head is a two-layer MLP whose hidden and output dimensions are 2048 and 128. The network uses the learned parameters $\theta$ for both views. 

\vspace{0.5em} \noindent \textbf{DetCon$_{B}$.} Following BYOL, the projection head is a two-layer MLP whose hidden and output dimensions are 4096 and 256. The network uses the learned parameters $\theta$ for processing one view, and an exponential moving average of these parameters $\xi$ for processing the second. Specifically, $\xi$ is updated using $\xi \leftarrow \lambda \cdot \xi + (1 - \lambda) \cdot \theta$, where the decay rate $\lambda$ is annealed over the course of training from $\lambda_0$ to 1 using a cosine schedule \cite{grill2020bootstrap}. $\lambda_0$ is set to 0.996 when training for 1000 epochs and 0.99 when training for 300 epochs. The projection of the first view is further transformed with a prediction head, whose architecture is identical to that of the projection head. 

\vspace{0.5em} \noindent \textbf{Computational cost.} The forward pass through a ResNet-50 encoder requires roughly 4B FLOPS. Ignoring the cost of bias terms and point-wise nonlinearities, the projection head in \dcs requires 4.4M FLOPS (i.e. 2048\x2048 $+$ 2048\x128). Since this is calculated 16 times rather than once, it results in an overhead of 67M FLOPS compared to SimCLR. For \dcb the combined cost of evaluating the projection and prediction heads results in an additional 173M FLOPS compared to BYOL. Finally, the cost of evaluating the contrastive loss is 134M FLOPS for \dcs (i.e.\ 128\x4096\x16$^2$) and 268M FLOPS for DetCon$_{B}$. In total \dcs requires 201M additional FLOPS and \dcb 
441M which represent 5.3\% and 11.6\% of the cost of evaluating the backbone. This overhead is sufficiently small compared to the gain in training iterations required to reach a given transfer performance (e.g.\ a 500\% gain for \dcs over SimCLR, and a 333\% for \dcb over DetCon) for us not further distinguish between gains in computation and training time.

\subsection{Implementation: optimization}
\label{sec:app-optimization}

\vspace{0.5em} \noindent \textbf{Self-supervised pretraining.} We train using the LARS optimizer \cite{you2017large} with a batch size of 4096 split across 128 Cloud TPU v3 workers. When training on ImageNet we again adopt the optimization details of SimCLR and BYOL for DetCon$_{S}$ and DetCon$_{B}$, scaling the learning rate linearly with the batch size and decaying it according to a cosine schedule. For \dcs the base learning rate is 0.3 and the weight decay is $10^{-6}$. \dcb also uses these values when training for 300 epochs; when training for 1000 epochs they are 0.2 and $1.5 \cdot 10^{-6}$.

When pretraining on COCO, we replace the cosine learning rate schedule with a piecewise constant, which has been found to alleviate overfitting \cite{he2019rethinking}, dropping the learning rate by a factor of 10 at the \nth{96} and \nth{98} percentiles. For fair comparison we use the same schedules when applying SimCLR to the COCO dataset, which we also find to perform better than the more aggressive cosine schedule.

\vspace{0.5em} \noindent \textbf{Transfer to COCO.} We fine-tune with stochastic gradient descent, increasing the learning rate linearly for the first 500 iterations and dropping twice by a factor of 10, after $\frac{2}{3}$ and $\frac{8}{9}$ of the total training time, following \cite{wu2019detectron2}. We use a base learning rate of 0.3 for ResNet-50 models and 0.2 for larger ones, a momentum of 0.9, a weight decay of 4$\cdot$10$^{-5}$, and a batch size of 64 images split across 16 workers. 

\vspace{0.5em} \noindent \textbf{Transfer to PASCAL.} We fine-tune for 45 epochs with stochastic gradient descent, with a batch size of 16 and weight decay of $10^{-4}$. The learning rate is 0.02 and dropped by a factor of 10 at the \nth{70} and \nth{90} percentiles.

\vspace{0.5em} \noindent \textbf{Transfer to Cityscapes.} We fine-tune for 160 epochs with stochastic gradient descent and a Nesterov momentum of 0.9, using a batch size of 2 and weight decay of $10^{-4}$. The initial learning rate is 0.005 and dropped by a factor of 10 at the \nth{70} and \nth{90} percentiles.

\vspace{0.5em} \noindent \textbf{Transfer to NYU-Depth v2.} We fine-tune for 7500 steps with a batch size of 256, weight decay of 5$\cdot10^{-4}$, and a learning rate of 0.16 scaled linearly with the batch size \cite{grill2020bootstrap}.% using a constant learning rate of 0.16 (scaled linearly to account for the bigger batch size). We apply a weight decay of 0.0005.

% We train for 7500 steps with batch size 256, weight decay 0.0005, and learning rate 0.16 (scaled linearly from the setup of [83] to account for the bigger batch size).

\subsection{Results: larger models}
\label{sec:app-larger-models}

\noindent In Table \ref{tab:prior_art_r50} we compare to prior works on self-supervised learning which transfer to COCO. Here we provide additional comparisons which use larger models (ResNet-101, -152, and -200). We find DetCon to continue to outperform prior work in this higher capacity regime (Table \ref{tab:prior_art_big}). 

\begin{table}[t]
\subfloat[\textbf{ResNet-101} feature extractor]{
\small
\begin{tabularx}{\linewidth}{c *{4}{Y}}
 & \multicolumn{2}{c}{Fine-tune 1\x}  
 & \multicolumn{2}{c}{Fine-tune 2\x}
 \\
\cmidrule(lr){2-3} \cmidrule(l){4-5} 
method & \apbbox{~} & \apmask{~} & \apbbox{~} & \apmask{~} \\
\midrule
Supervised                     & 42.0 & 37.3 & 43.4 & 38.4 \\
SimCLR \cite{chen2020simple}   & 42.0 & 37.9 & 43.8 & 39.3 \\ 
InfoMin \cite{tian2020makes}   & 42.9 & 38.6 & 44.5 & 39.9 \\ 
BYOL \cite{grill2020bootstrap} & 43.7 & 38.8 & 44.3 & 39.4 \\ 
\midrule
\textbf{DetCon$_{B}$} & \textbf{45.2} & \textbf{40.0} & \textbf{45.7} & \textbf{40.4} \\ 
\end{tabularx}
}  % end of subfloat

\subfloat[\textbf{ResNet-152} feature extractor]{
\small
\begin{tabularx}{\linewidth}{c *{4}{Y}}
 & \multicolumn{2}{c}{Fine-tune 1\x}  
 & \multicolumn{2}{c}{Fine-tune 2\x}
 \\
\cmidrule(lr){2-3} \cmidrule(l){4-5} 
method & \apbbox{~} & \apmask{~} & \apbbox{~} & \apmask{~} \\
\midrule
Supervised                     & 43.4 & 38.5 & 43.4 & 38.5 \\
SimCLR \cite{chen2020simple}   & 43.6 & 39.1 & 44.9 & 40.0 \\
BYOL \cite{grill2020bootstrap} & 44.9 & 40.0 & 45.7 & 40.6 \\
\midrule
\textbf{DetCon$_{B}$}          & \textbf{46.0} & \textbf{40.6} & \textbf{46.4} & \textbf{40.7} \\
\end{tabularx}
}  % end of subfloat

\subfloat[\textbf{ResNet-200} feature extractor]{
\small
\begin{tabularx}{\linewidth}{c *{4}{Y}}
 & \multicolumn{2}{c}{Fine-tune 1\x}  
 & \multicolumn{2}{c}{Fine-tune 2\x}
 \\
\cmidrule(lr){2-3} \cmidrule(l){4-5} 
method & \apbbox{~} & \apmask{~} & \apbbox{~} & \apmask{~} \\
\midrule
Supervised                     & 43.2 & 38.3 & 43.5 & 38.5 \\
SimCLR \cite{chen2020simple}   & 44.3 & 39.6 & 45.3 & 40.3 \\ 
BYOL \cite{grill2020bootstrap} & 45.6 & 40.5 & 45.9 & 40.5 \\ 
\midrule
\textbf{DetCon$_{B}$}          & \textbf{47.1} & \textbf{41.3} & \textbf{47.2} & \textbf{41.5} \\
\end{tabularx}
}  % end of subfloat
\vspace{.5em}
\caption{\textbf{Comparison to prior art:} all methods are pretrained on ImageNet then fined-tuned on COCO for 12 epochs (1\x ~ schedule) or 24 epochs (2\x ~ schedule). %Bounding-box AP (\apbbox{}) and mask AP (\apmask{}) are evaluated on the COCO \texttt{val2017} set.
}
\label{tab:prior_art_big}
\vspace{-1.em}
\end{table}

% \subsection{Errata}
% \label{sec:errata}

% Our initial submission contains two minor mistakes which we correct here. 

% \vspace{0.5em} \noindent \textbf{Computational complexity.} Our initial submission claimed that the computational complexity of the contrastive loss represents 0.1\% of the cost of evaluating a ResNet-50 backbone. This is incorrect. As we describe in Section \ref{sec:app-architecture}, the contrastive loss requires 134M FLOPS for \dcs and 268M FLOPS for \dcb which make up 3.5\% and 7.1\% of cost of evaluating the backbone. In total, forward passes for \dcs and \dcb are 5.3\% and 11.6\% more expensive than those in SimCLR and BYOL. As a result, our conclusion remains that their cost is sufficiently similar (compared to the large gains in the number of training iterations) for us to equate gains in the number of training iterations with gains in total computational cost.

% \vspace{0.5em} \noindent \textbf{References.} Our initial submission mistakenly truncated the list of references to the first 59. We include the full reference list here. 

{\small
\bibliographystyle{ieee_fullname}
\bibliography{egbib}
}

\end{document}